\documentclass[10pt,twocolumn,letterpaper]{article}

\usepackage{cvpr}
\usepackage{times}
\usepackage{epsfig}
\usepackage{graphicx}
\usepackage{amsmath}
\usepackage{amssymb}

\usepackage{color}
\usepackage{booktabs}
\usepackage{multirow}

\usepackage[pagebackref=true,breaklinks=true,letterpaper=true,colorlinks,bookmarks=false]{hyperref}

\cvprfinalcopy 

\def\cvprPaperID{1151} 

\ifcvprfinal\fi
\begin{document}

\title{Information Maximizing Visual Question Generation}

\author{Ranjay Krishna, Michael Bernstein, Li Fei-Fei\\
Stanford University\\
{\tt\small \{ranjaykrishna, msb, feifeili\}@stanford.edu}
}

\maketitle

\begin{abstract}
Though image-to-sequence generation models have become overwhelmingly popular in human-computer communications, they suffer from strongly favoring safe generic questions (``What is in this picture?''). Generating uninformative but relevant questions is not sufficient or useful. We argue that a good question is one that has a tightly focused purpose --- one that is aimed at expecting a specific type of response. We build a model that maximizes mutual information between the image, the expected answer and the  generated question. To overcome the non-differentiability of discrete natural language tokens, we introduce a variational continuous latent space onto which the expected answers project. We regularize this latent space with a second latent space that ensures clustering of similar answers. Even when we don't know the expected answer, this second latent space can generate goal-driven questions specifically aimed at extracting objects (``what is the person throwing''),  attributes, (``What kind of shirt is the person wearing?''), color (``what color is the frisbee?''), material (``What material is the frisbee?''), etc. We quantitatively show that our model is able to retain information about an expected answer category, resulting in more diverse, goal-driven questions. We launch our model on a set of real world images and extract previously unseen visual concepts.
\end{abstract}

\section{Introduction}
The task of transforming visual scenes to language, questions~\cite{ren2015exploring,mostafazadeh2016generating}, answers~\cite{zhu2016visual7w,antol2015vqa} or captions~\cite{vinyals2015show,krause2017hierarchical}, has widely adopted an image-to-sequence architecture that encodes the image through a convolutional neural network (CNN)~\cite{krizhevsky2012imagenet} and then decodes the language with a recurrent neural network~\cite{mikolov2010recurrent}. The whole framework can be efficiently trained by maximum likelihood estimation (MLE) and has demonstrated state-of-the-art performance in various tasks~\cite{xu2015show,cho2015describing,das2017human,donahue2015long,lu2018neural}. However, this training procedure is not suitable for generating questions or enabling discovery of new concepts. In fact, most MLE-based training schemes have shown to produce generic questions that result in uninformative answers (e.g.~``yes'')~\cite{jabri2016revisiting}, questions (e.g.~``What is the person doing?'')~\cite{jain2017creativity}, captions (e.g.~``A clear day with a blue sky''~\cite{xu2015show}) or dialogue (e.g.~``I don't know'')~\cite{li2015diversity,serban2016building}. Simply generating a generic question is not sufficient or useful for discovering new concepts.

\begin{figure}[t]
    \centering
    \includegraphics[width=\columnwidth]{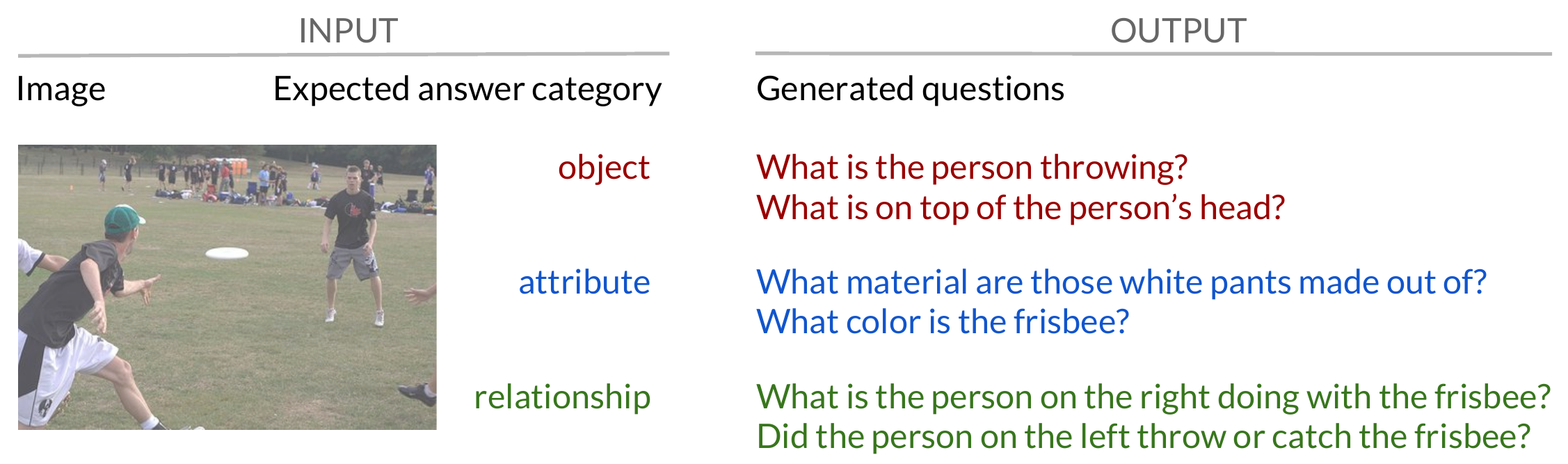}
\caption{Our new architecture generates goal-driven visual questions that maximize the likelihood of receiving an expected answer. When attempting to learn about objects or their attributes, it can generate questions aimed at attaining such answer categories.}
    \label{fig:pull_figure}
\end{figure}

\begin{figure*}[t]
    \centering
    \includegraphics[width=\linewidth]{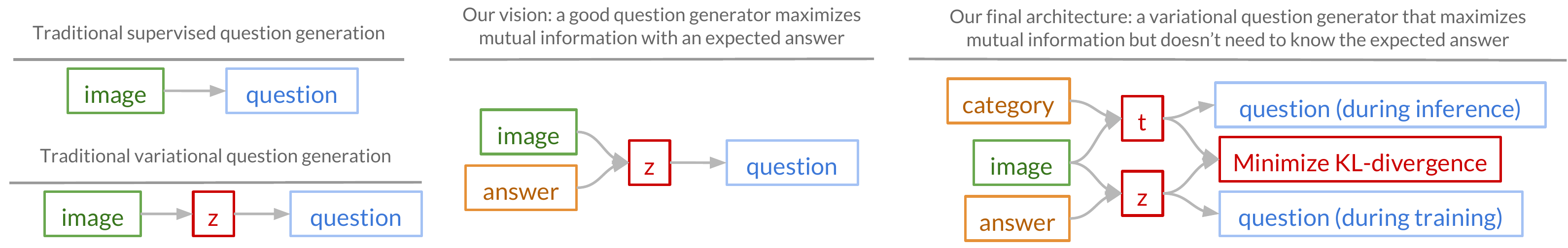}
\caption{Since multiple questions are possible for any image, previous approaches moved away from supervised question generation to variational approaches~\cite{jain2017creativity}. However, this resulted in generic, uninformative questions. We argue that a good question maximizes mutual information with an expected answer. But such a model is not practical as knowing the answer defeats the purpose of generating a question. Also, such models often lead to the posterior collapsing problem~\cite{bowman2015generating}. Instead, we propose an architecture that maximizes mutual information between the image, answer and question while also maintaining a regularization based on answer categories. Our final model uses two latent spaces and can generate questions both in the presence and the absence of answers.}
    \label{fig:high_level}
\end{figure*}

Instead of generating generic questions, question generation models should be goal-driven --- we show how they can be trained to ask questions aimed at extracting specific answer categories. Visual question generation is not a bijection, i.e.~multiple correct questions can be generated from the same image. Previous research moved away from a supervised approach to question generation to a variational approach that can generate multiple questions by sampling a latent space~\cite{jain2017creativity} (see Figure~\ref{fig:high_level}). However, previous approaches are not goal-driven --- they do not guarantee that the question will result in a specific type of answer. To remedy their problem, we could encode the answer along with the image before generating the question. While such an approach allows the model to condition its question on the answer, it is neither technically feasible nor practical. Technical infeasibility arises because variational models often lead to the posterior collapsing problem~\cite{bowman2015generating}, where the model can learn to ignore the answer when generating questions. Impracticality arises because the main purpose of asking questions is to attain an answer, implying that knowing the answer defeats the purpose of generating the question. 

To tackle the first challenge, we design a visual question generation architecture that maximizes the mutual information between the generated question with the image as well as with the expected answer (see Figure~\ref{fig:high_level}). We call our model \textbf{Information Maximizing Visual Question Generator} as it maximizes relevance with the image and expectation over the answer. Safe, generic questions that lead to uninformative answers are discouraged as they have low mutual information with either. However, optimizing for mutual information is often intractable and given the discrete tokens (words) we wish to generate, no unbiased, low variance gradient estimator exists~\cite{jang2016categorical,maddison2016concrete,he2016dual,shetty2017speaking,paulus2017deep,williams1992simple,kaiser2018fast}. We formulate our model as a variational auto-encoder that attempts to learn a joint continuous latent space between the image, question and the expected answer. Instead of directly optimizing discrete utterances, the question, image and expected answer are all trained to maximize the mutual information with this latent space. By reconstructing the image and expected answer representations, we can maximize the evidence lower bound (ELBO) and control what information the generated questions request.  

The second challenge arises from the lack of an expected answer in real world deployments. Since we require an answer to map the image into a latent space, it is not possible to generate questions in the absence of an answer. Enumerating all possible answers is infeasible. Instead, we propose creating a second latent space that is learned from the image and the answer category instead of the answer itself. Answer categories can be objects, attributes, colors, materials time, etc. During training, we minimize the KL-divergence between these two latent spaces. Not only does this allow us to generate visual questions that maximize mutual information with the expected answer, it also acts as a regularizer into the original latent space. It prevents the learned latent spaces from overfitting to specific answers in the training set and forces them to generalize to categories of questions.

We annotate the VQA dataset~\cite{antol2015vqa} with $15$ categories for the top $500$ answers and use it to train our model, which queries for specific answer categories. We evaluate our model on relevance to the image and on its ability to expect the answer type. Finally, we run our model on $1000$ real world images and discover $80$ new objects, $40$ new attributes, $17$ new colors, and $8$ new materials.

\section{Related work}
Visual understanding has been studied vigorously through question answering with the availability of large scale \textbf{visual question answering} (VQA) datasets~\cite{antol2015vqa,zhu2016visual7w,krishna2017visual,johnson2017clevr}. Current VQA approaches follow a traditional supervised MLE paradigm that typically relies on a CNN + RNN encoder-decoder formulation~\cite{vinyals2015show}. Successive models have improved performance by stacking attention~\cite{yang2016stacked,lu2016hierarchical}, modularizing components~\cite{andreas2016learning,johnson2017inferring,hu2017learning}, adding relation networks~\cite{santoro2017simple}, augmenting memory~\cite{xiong2016dynamic}, and adding proxy tasks~\cite{fukui2016multimodal,wang2017joint}. While the performance of VQA models have been encouraging, they require a large labelled dataset with a predefined vocabulary. In contrast, we focus on the surrogate task of generating questions in the hopes of augmenting real world agents with the ability to expand it's visual knowledge by discovering new visual concepts.

\begin{figure*}[t!]
    \centering
    \includegraphics[width=\linewidth]{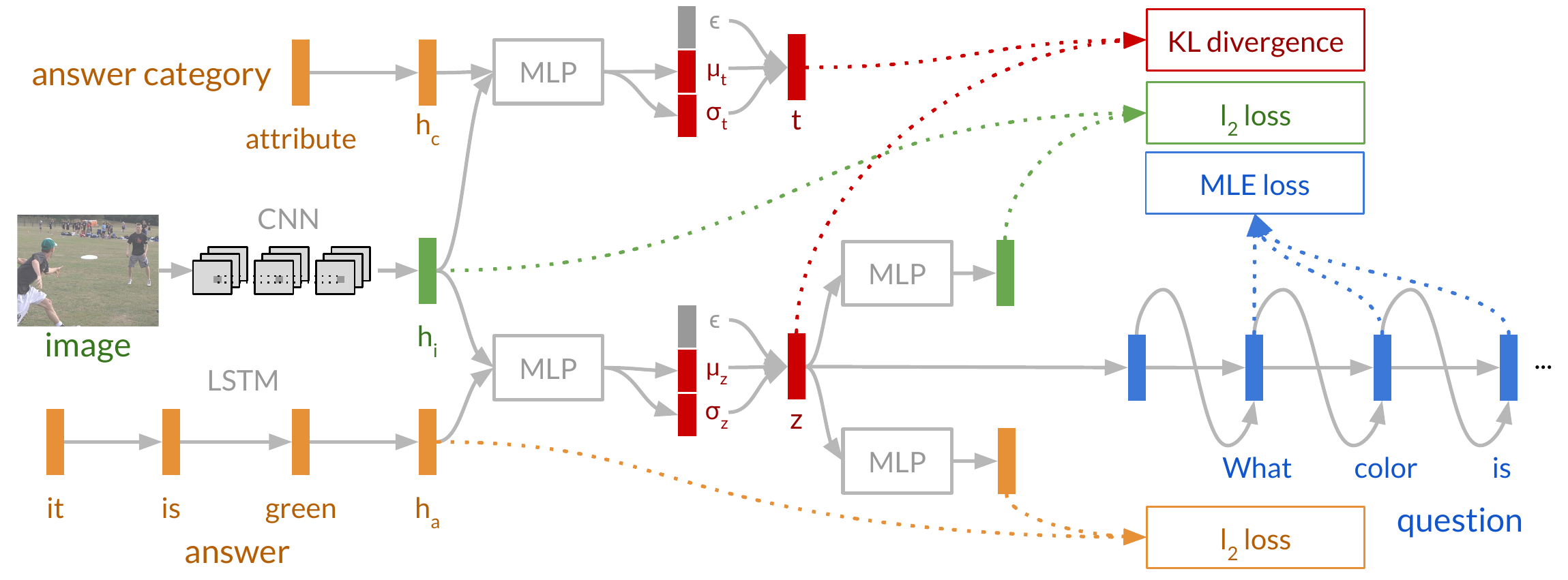}
    \caption{Training our model: we embed the image and answer into a latent space $z$ and attempt to reconstruct them, thereby maximizing mutual information with the image and the answer. We also use $z$ to generate questions and train it with an MLE objective. Finally, we introduce a second latent space $t$ that is trained by minimizing KL-divergence with $z$. $t$ allows us to remove the dependence on the answer when generating questions and instead grants us the ability to generate questions conditioned on the answer category.}
    \label{fig:model}
\end{figure*}

In contrast to answering questions, generating questions has received little interest so far. In NLP, a few methods have attempted to automatically generate questions from knowledge bases using rule based~\cite{serban2016generating} or deep learning based systems~\cite{du2017learning}. In computer vision, a few recent projects have explored the task of \textbf{visual question generation} to build curious visual agents~\cite{yang2018visual,jain2017creativity}. These projects have also either followed an algorithmic rule-based~\cite{vijayakumar2016diverse,serban2016generating} or learning-based~\cite{mostafazadeh2016generating,ren2015exploring} approach. Newer papers have treated the generation process as a variational process~\cite{jain2017creativity} or placed it in a active learning~\cite{misra2017learning} or reinforcement learning setting~\cite{yang2018visual}. Our work draws inspiration from these previous methods and extends them by treating question generation as a process that maximizes mutual information between not just the image but also considers the expected answer's category. We believe that a good question generator should be goal driven --- it should generate questions to receive a particular answer category.

There is a large body of work exploring \textbf{generative models} and learning latent representation spaces. Early work focused primarily on stacked autoencoders and then on restricted boltzman machines~\cite{vincent2008extracting,hinton2006fast,hinton2006reducing}. Recent successes of these applications have primarily been a result of variational auto-encoders (VAEs)~\cite{kingma2013auto} and generative adversarial networks (GANs)~\cite{goodfellow2014generative}. With the reparameterization trick, VAEs can be trained to learn a semi-supervised latent space to generate images~\cite{kingma2013auto}. They have also been extended to continuous state space~\cite{krishnan2015deep,archer2015black} and sequential models~\cite{gregor2015draw,chung2015recurrent}. GANs, on the other hand, can learn image representations that support basic linear algebra~\cite{radford2015unsupervised} and even enable one-shot learning by using probabilistic inference over Bayesian programs~\cite{lake2015human}. Both VAEs and GANs have disentangled their representations based on class labels or other visual variations~\cite{kingma2014semi,makhzani2015adversarial}. While we do not explicitly disentangle the representation, we will demonstrate later how the second latent space regularizes the original space and disentangles the representations of different answer categories. 

Generative models often require a series of tricks for successful training~\cite{salimans2016improved,radford2015unsupervised,burda2015importance,bowman2015generating}. And even with these tricks, training them with discrete tokens is only possible by using \textbf{gradient estimators}. As we mentioned earlier, these estimators often suffer from one of two problems: high bias~\cite{kaiser2018fast,kaiser2018fast} or high variance~\cite{williams1992simple}. Low variance methods like Gumbel-Softmax~\cite{jang2016categorical}, CONCRETE distribution~\cite{maddison2016concrete}, semantic hashing~\cite{kaiser2018fast} or vector quantization~\cite{van2017neural} result in biased estimators. Similarly, low bias methods like REINFORCE~\cite{williams1992simple} with Monte Carlo rollouts, result in high variance~\cite{he2016dual,shetty2017speaking,paulus2017deep}. We overcome this issue by introducing a continuous latent space that maximizes mutual information with encodings of the image, question and answer. This latent space can be trained using existing VAE training procedures that attempt to reconstruct the image and answer representations. We further extend this model with a second latent space conditioned on the answer category that removes the need for an actual answer when generating questions.

\section{Information Maximizing Visual Question Generator}

\begin{figure*}[t]
    \centering
    \includegraphics[width=0.8\linewidth]{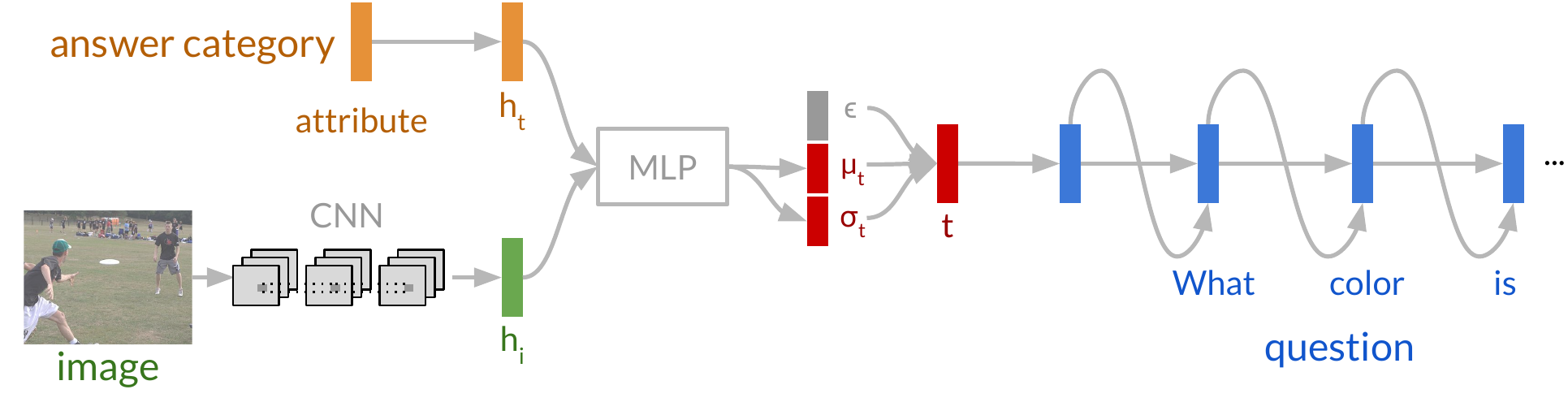}
    \caption{Inference on our model: Given an image input and an answer category (e.g.~attribute), we encode both into a latent representation $t$, parameterized by mean $\mu_t$ and $\sigma_t$. We sample from $t$ with noise $\epsilon$ to generate questions that are relevant to the image and who's answers result in the given answer category.}
    \label{fig:inference}
\end{figure*}

Our aim is to generate questions that have a tightly focused purpose --- questions with the aim to learn something specific about the image. Agents with the capability to request specific categories of information can extract new concepts more effectively from the real world. In this section, we detail how we design an Information Maximizing Visual Question Generator. Recall that the goal of our model is to generate questions given an image and an answer category. For example, if we want to understand materials or binary answers, our model should generate questions ``What material is that desk made out of?'' or ``Is the desk on the right of the chair?'', respectively. Our two challenges are (1) technical infeasibility caused by non-differentiable discrete tokens and variational posterior collapse and (2) impracticality of requiring answers to generate questions. We start off with a formal definition of the problem, explain why current methods fail and then detail our training and inference process.

\subsection{Problem formulation}
Let $q$ denote the question we want to generate for an image $i$. This question should result in the an answer $a$ of category $c$. For example, the question ``What is the person in red doing with the ball?'' should result in the answer ``kicking'', which belong to category ``activity''. Our final goal is to define a model $p(q \vert i, c)$. But first, let's attempt to define a simpler model $p(q \vert i, a)$ that maximizes the mutual information between the image and the question $I(i, q)$ and between the expected answer and the question $I(a, q)$. This objective can be written as:
\begin{equation}
\begin{aligned}
    \max I(i, q) + \lambda I(a, q)\\
    \textrm{s. t.  } q \sim p(q \vert i, a)
\end{aligned}
\end{equation}
where $\lambda$ is a hyperparameter that adjusts for their relative importance in the optimization.

\subsection{Continuous latent space}
As already mentioned, directly optimizing this objective is infeasible because the exact computation of mutual information is intractable. Additionally, optimizing by estimating gradients between discrete steps is difficult as the estimator needs to have both low bias and low variance. To overcome this challenge, we introduce a continuous, dense, latent $z$-space. We learn a mapping $p_\theta(z \vert i, a)$, parameterized by $\theta$, from the image and the expected answer to this latent space.

With this $z$-space, our new optimization becomes:
\begin{equation}
\begin{aligned}
    \max_\theta I(q, z \vert a, i) + \lambda_1 I(a, z) + \lambda_2 I(i, z)\\
    \textrm{s. t.  } z \sim p_\theta(z \vert i, a)\\
    \textrm{       } q \sim p_\theta(q \vert z)
\end{aligned}
\label{eq:cont}
\end{equation}
where $\lambda_1$ and $\lambda_2$ are hyperparameters that relatively weight the mutual information terms in the optimization.

\subsection{Variational mutual information maximization}
So far, we have avoided discrete tokens. However, this mutual information maximization is still intractable as it requires knowing the true posteriors $p(z \vert i)$ and $p(z \vert a)$. Fortunately, we can opt to maximize its ELBO:
\begin{equation}
\begin{aligned}
    I(z, i) &= \mathop{\mathbb{H}}(i) - \mathop{\mathbb{H}}(i \vert z)\\
    &=\mathop{\mathbb{H}}(i) + \mathop{\mathbb{E}}{}_{z \sim p(z, i)}[\mathop{\mathbb{E}}{}_{\hat{i} \sim p(i \vert z)}[\log p(\hat{i} \vert z)]]\\
    &=\mathop{\mathbb{H}}(i) + \mathop{\mathbb{E}}{}_{i \sim p(i)} [D_{KL}[p(\hat{i} \vert z) \vert\vert p_\theta (\hat{i} \vert z)] \\
    &\;\;\;\; + \mathop{\mathbb{E}}{}_{\hat{i} \sim p(i \vert z)}[\log p_\theta (\hat{i} \vert z)]]\\
    &\geq \mathop{\mathbb{H}}(i) + \mathop{\mathbb{E}}{}_{i \sim p(i)} [\mathop{\mathbb{E}}{}_{\hat{i} \sim p(i \vert z)}[\log p_\theta (\hat{i} \vert z)]]]
\end{aligned}
\label{eq:i_z_i}
\end{equation}
where $\mathop{\mathbb{H}}(\cdot)$ is the entropy function and $\mathop{\mathbb{E}}$ is expectation. $p_\theta(\cdot)$ is a function parameterized by $\theta$. This optimization is often referred to as variational information maximization~\cite{chen2016infogan}. Similarly,
\begin{equation}
\begin{aligned}
I(z, a) \geq \mathop{\mathbb{H}}(a) + \mathop{\mathbb{E}}{}_{a \sim p(a)} [\mathop{\mathbb{E}}{}_{\hat{a} \sim p(a \vert z)}[\log p_\theta (\hat{a} \vert z)]]] 
\end{aligned}
\label{eq:i_z_a}
\end{equation}

The third and final conditional mutual information term $I(q, z \vert a, i)$ can also be bounded by:
\begin{equation}
\begin{aligned}
I(z, q \vert a, i) &\geq \mathop{\mathbb{H}}(q) + \\
&\mathop{\mathbb{E}}{}_{q \sim p(q \vert i, a)} [\mathop{\mathbb{E}}{}_{\hat{q} \sim p(q \vert z, a, i)}[\log p_\theta (\hat{q} \vert z, i, a)]]] \\
&\textrm{s. t.  } p(q \vert z, a, i) = p(q \vert z)p(z \vert a, i)
\end{aligned}
\label{eq:i_z_q}
\end{equation}

Putting  Eq.~\ref{eq:i_z_i},~\ref{eq:i_z_a} and~\ref{eq:i_z_q} together in Eq.~\ref{eq:cont}:
\begin{equation}
\begin{aligned}
\max_\theta \mathop{\mathbb{E}}{}_{p_\theta (q,i,a)} [&\log p_\theta (q \vert i,a, z) + \lambda_1 \log p_\theta (a \vert z) \\&+ \lambda_2 \log p_\theta (i \vert z)]\\
    \textrm{s. t.  } &p_\theta (q,i,a) = p_\theta (q \vert z) p_\theta (z \vert i,a) p(i,a)
\end{aligned}
\label{eq:reconstruction}
\end{equation}
Note that we ignore the entropy terms associated with the training data as it doesn't involve the parameter $\theta$ we are trying to optimize. Therefore, optimizing Eq.~\ref{eq:reconstruction} can be accomplished by maximizing the reconstruction of the image and answer representations while maximizing the MLE objective of generating the question.

\subsection{Question generation by reconstructing image and answer representations}
To functionalize the optimization presented above, we begin by first encoding the image using a CNN as a dense vector $h_i$ (see Figure~\ref{fig:model}). Similarly, we encode the answer $a$ using a long short term memory network (LSTM)~\cite{hochreiter1997long}, which is a variant of RNNs, into another dense vector $h_a$. Next, we feed $h_i$ and $h_a$ into a VAE that embeds both into a latent $z$-space. In practice, we assume that $z$ follows a multivariate Gaussian distribution with diagonal covariance. We use the reparameterization trick~\cite{kingma2013auto}, to generate means $\mu_z$ and standard deviations $\sigma_z$, combine it with a sampled unit Gaussian noise $\epsilon$ to generate $z = \mu_z + \sigma_z \epsilon$. 

From $z$, we reconstruct $\hat{h}_i$ and $\hat{h}_a$ and optimize the first two terms in Eq.~\ref{eq:reconstruction} by minimizing the following $l_2$ losses:
\begin{equation}
\begin{aligned}
L_i = \lvert \vert h_i - \hat{h}_i \vert \rvert_2, \;\;\;\; L_a = \lvert \vert h_a - \hat{h}_a \vert \rvert_2
\end{aligned}
\end{equation}
Next, we use a decoder LSTM to generate the question $\hat{q}$ from $z$-space. We minimize the MLE objective $L_{MLE}$ between $\hat{q}$ and the true question in our training set $q$, which results in the third and final term in Eq.~\ref{eq:reconstruction}.

\subsection{Regularizing with a second latent space}
So far, we have proposed building a model that maximizes the lower bound of mutual information between a latent space, the image and the expected answer. This allows us to generate questions if we know what the expected answer should be. This is not conducive to our original goal of deploying our model in real world situations where it does not know the answer a priori. If we already know the answer to a question, there is no point in generating a question.

To remedy this, we propose a second latent $t$-space. Instead of using both $a$ and $i$ to encode $h_a$ and $h_i$ into $z$-space, we discard the answer and instead only use its category $c$. We classify answers as being one of a few predefined categories, such as objects (e.g.~``cat''), attributes (e.g.~``cold''), color (e.g.~``brown''), relationship (e.g.~``ride''), counting (e.g.~``1''), etc. These categories are cast as a one hot vector and encoded as $h_c$ and used, along with $h_i$ to embed into the variational $t$-space. We train $t$-space by minimizing the KL-divergence with $z$-space:
\begin{equation}
\begin{aligned}
    L_{t} &= D_{KL}(p_\theta(z\vert i, a), p_\phi(t \vert i, c))\\
    &= \log{\sigma_t} - \log{\sigma_z} + \frac{\sigma_z + (\mu_t - \mu_z)^2}{2\sigma_t}- 0.5
\end{aligned}
\end{equation}
where $\phi$ are the parameters used to embed into $t$-space. This allows us to now utilize $p_\phi(t \vert i, c)$ to embed into a space that closely resembles $z$-space. Since we assume that both $z$-space and $t$-space follow a multivariate Gaussian with diagonal covariance, the KL term has the analytical form shown above. We no longer need to know the answer $a$ to embed and generate questions. Intuitively, the $t$-space can be also thought of as a regularizer on $z$-space, preventing the model from overfitting to the answers in the training data and relying instead on utilizing the answer categories.

Putting them together, the final loss for our model is:
\begin{equation}
    L = L_{MLE} + \lambda_1 L_a + \lambda_2 L_i + \lambda_3 L_t
\end{equation}
where $\lambda_1$ and $\lambda_2$ have already been introduced and $\lambda_3$ is a hyperparameter that controls the amount of regularization used in our model. Note that we are omitting the KL-loss with respect to a unit normal centered at zero that maintains the two latent spaces' priors.

\subsection{Inference}
During inference, we are given an image $i$ and answer category $c$ and are expected to generate questions. We encode the inputs into the second latent $t$-space and sample from it to generate questions, as shown in Figure~\ref{fig:inference}. This allows us to generate goal-driven questions for any image, focused towards extracting its objects, its attributes, etc.

\subsection{Implementation details}
We implement our model using PyTorch and plan on releasing all our code. We use ResNet18~\cite{he2016deep} as our image encoder and do not fine-tune its weights. $h_i$, $h_a$ and $h_t$ are all $512$ dimensional vectors. $z$-space and $t$-space are $100$ dimensions. The encoders for the image and answer are trained only from $L_{MLE}$ and not $L_i$, $L_a$ or $L_t$ to prevent the encoders from simply optimizing for the reconstruction loss at the cost of not being able to generate questions. We optimized the hyperparameters such that $\lambda_1 = 0.01$, $\lambda_2 = 0.001$, $\lambda_3 = 0.005$ with a learning rate of $0.001$ that decays every $4$ epochs for a total of $10$ epochs.

\section{Experiments}

\begin{table*}[t]
\centering
\small
\caption{We report our model's efficacy with multiple metrics. We use language modeling metrics to measure its capability to generate questions similar to the ground truth. Next, we measure the model's ability to maximize mutual information by predicting the answer or its category from the latent space embedding. Finally, we measure the relevance of the question with the image. Note that language modeling scores are multiplied by $100$ to show more significant digits and mutual information and relevance scores are reported in percentages.}
\label{tab:mi}
\begin{tabular}{p{.1cm}p{1.7cm} p{0.9cm}p{0.9cm}p{0.9cm}p{0.9cm} p{1.0cm}p{1.0cm}     p{0.1cm}p{1.0cm}p{1.0cm}    p{0.1cm}p{1.0cm}p{1.0cm}}
    & & \multicolumn{6}{c}{Language modeling} & & \multicolumn{2}{c}{Mutual information} & & \multicolumn{2}{c}{Relevance}\\
    \cmidrule{3-8}\cmidrule{10-11}\cmidrule{13-14}
    & Models & Bleu-1 & Bleu-2 & Bleu-3 & Bleu-4 & METEOR & CIDEr & & Answer & Category & & Image & Category\\
    \midrule\midrule
    \multirow{6}{*}{\rotatebox[origin=c]{90}{$z$-space}} 
    & IA2Q~\cite{wang2017joint} & 32.43 & 15.49 & 9.24 & 6.23 & 11.21 & 36.22 & & 11.48 & 35.33 & & 91.10 & 36.80 \\
    & V-IA2Q~\cite{jain2017creativity} & 36.91 & 17.79 & 10.21 & 6.25 & 12.39 & 36.39 & & 11.13 & 36.91 & & 90.10 & 39.00 \\
    & Ours w/o A & 38.88 & 20.74 & 12.75 & 6.29 & 12.78 & 40.13 & & 10.02 & 40.44 & & 98.10 & 42.70\\
    & Ours w/o AC & 38.99 & 21.48 & 12.73 & 6.57 & 13.01 & 42.13 & & 10.10 & 60.00 & & 96.80 & 42.80 \\
    & Ours w/o C & \textbf{50.09} & \textbf{32.32} & \textbf{24.61} & \textbf{16.27} & \textbf{20.58} & \textbf{94.33} & & \textbf{33.44} & 61.04 & & \textbf{98.00} & 82.40 \\
    & Ours & 48.09 & 29.76 & 20.71 & 15.17 & 18.78 & 92.13 & & 30.23 & \textbf{91.02} & & 97.10 & \textbf{91.20} \\
    \midrule
    \multirow{4}{*}{\rotatebox[origin=c]{90}{$t$-space}} 
    & IC2Q & 30.42 & 13.55 & 6.23 & 4.44 & 9.42 & 27.42 & & 9.88 & 40.23 & & 90.00 & 38.80 \\
    & V-IC2Q & 35.40 & 25.55 & 14.94 & 10.78 & 13.35 & 42.54 & & 10.11 & 60.23 & & 92.20 & 45.00 \\
    & Ours w/o A & 31.20 & 16.20 & 11.18 & 6.24 & 12.11 & 35.89 & & 9.35 & 68.23 & & \textbf{98.00} & 52.50\\
    & Ours  & \textbf{47.40} & \textbf{28.95} & \textbf{19.93} & \textbf{14.49} & \textbf{18.35} & \textbf{85.99} & & \textbf{28.23} & \textbf{99.02} & & 97.20 & \textbf{98.00}
    \end{tabular}
\end{table*}

\begin{figure*}[t]
    \centering
    \includegraphics[width=0.7\linewidth]{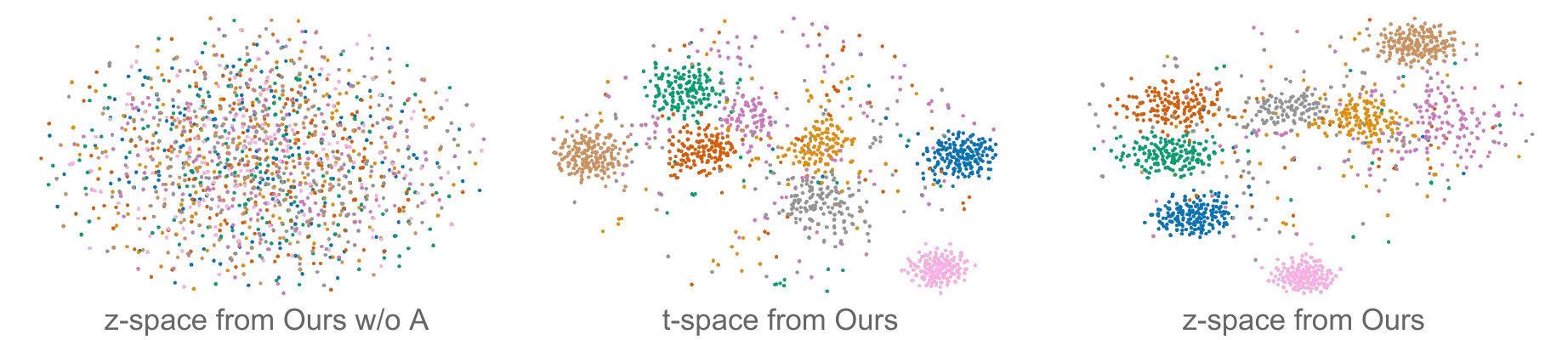}
\caption{TSNE~\cite{maaten2008visualizing} visualization of the latent encodings. When we don't reconstruct the answer, the embedding show no separation between answers or their categories, confirming the posterior collapse. Meanwhile, by reconstructing the answer, both the $z$-space and the $t$-space encodings are visually separable. Different colors represent categories of answers and we only show $8$ categories for aesthetics.}
    \label{fig:tsne}
\end{figure*}

To test our visual question generation model, we perform a series of experiments and evaluate the model along multiple dimensions. We start by discussing the dataset and evaluation metrics used. We then showcase examples of our model's generated questions when conditioned on the answer. Next, we demonstrate its ability when conditioned only on the answer category. We compare both these cases against a series of baselines and ablations. We analyze the diversity of questions produced within each answer category. Finally, we report a small proof of concept deployment of our model on real world images found online and show that it can learn new concepts.

\subsection{Experimental setup}
\noindent\textbf{Dataset.} To enable the kind of interaction where we can specify input answer categories, we need a VQA dataset that categorizes its answers. The VQA dataset~\cite{antol2015vqa} has a few basic categorizations of questions but not their answers. We annotate the VQA~\cite{antol2015vqa} dataset answers with a set of $15$ categories and label their top $500$ answers. These categories include objects (e.g.~``cat'', ``person''), attributes (e.g.~``cold'', ``old''), color (e.g.~``brown'', ``red''), relationship (e.g.~``ride'', ``jump''), counting (e.g.~``1'', ``10''), etc. The top $500$ answers make up the $82\%$ of the VQA dataset, resulting in $367K$ training+validation examples. We treat their validation set as our test set as the answers in their test set are not publicly available. We break the training set up into a $80$-$20\%$ train-validation split.

\noindent\textbf{Evaluation metrics.}
All past question generation papers have used a variety of evaluation metrics to calculate the quality of a question. While some have focused on maximizing diversity~\cite{vijayakumar2016diverse,jain2017creativity,zhang2016automatic}, others have treated it as a proxy task to improve question answering~\cite{li2018visual,ren2015exploring,wang2017joint}. Diversity measures have included using variants of beam search~\cite{vijayakumar2016diverse}, measuring novel questions or unique tri-grams~\cite{jain2017creativity} or creating rule-based datasets~\cite{zhang2016automatic}. Proxy tasks have typically used accuracy of multiple-choice answers to measure the performance of question generation.

We too report a variety of different evaluation metrics to highlight different components of our model. First, we use \textit{language modeling} evaluation metrics like BLEU, METEOR and CIDEr~\cite{chen2015microsoft} to calculate how well our generated questions match the ground truth questions in our test set. Next, we measure the \textit{mutual information} retained in the latent space by training a classifier to classify answer categories encoded in the latent space. This metric sheds light on how well our method retains information about the input answers or answer categories. Next, we measure \textit{relevance} of the question, ensuring that the questions are valid for the given image and result in the expected answer category. Relevance results are calculated from majority vote conducted by hiring $3$ crowd-workers that vote on whether a question can be answered given its corresponding image. Finally, we report diversity scores for each category, which measures the number of unique questions generated.

\begin{figure*}[t]
    \centering
    \includegraphics[width=\linewidth]{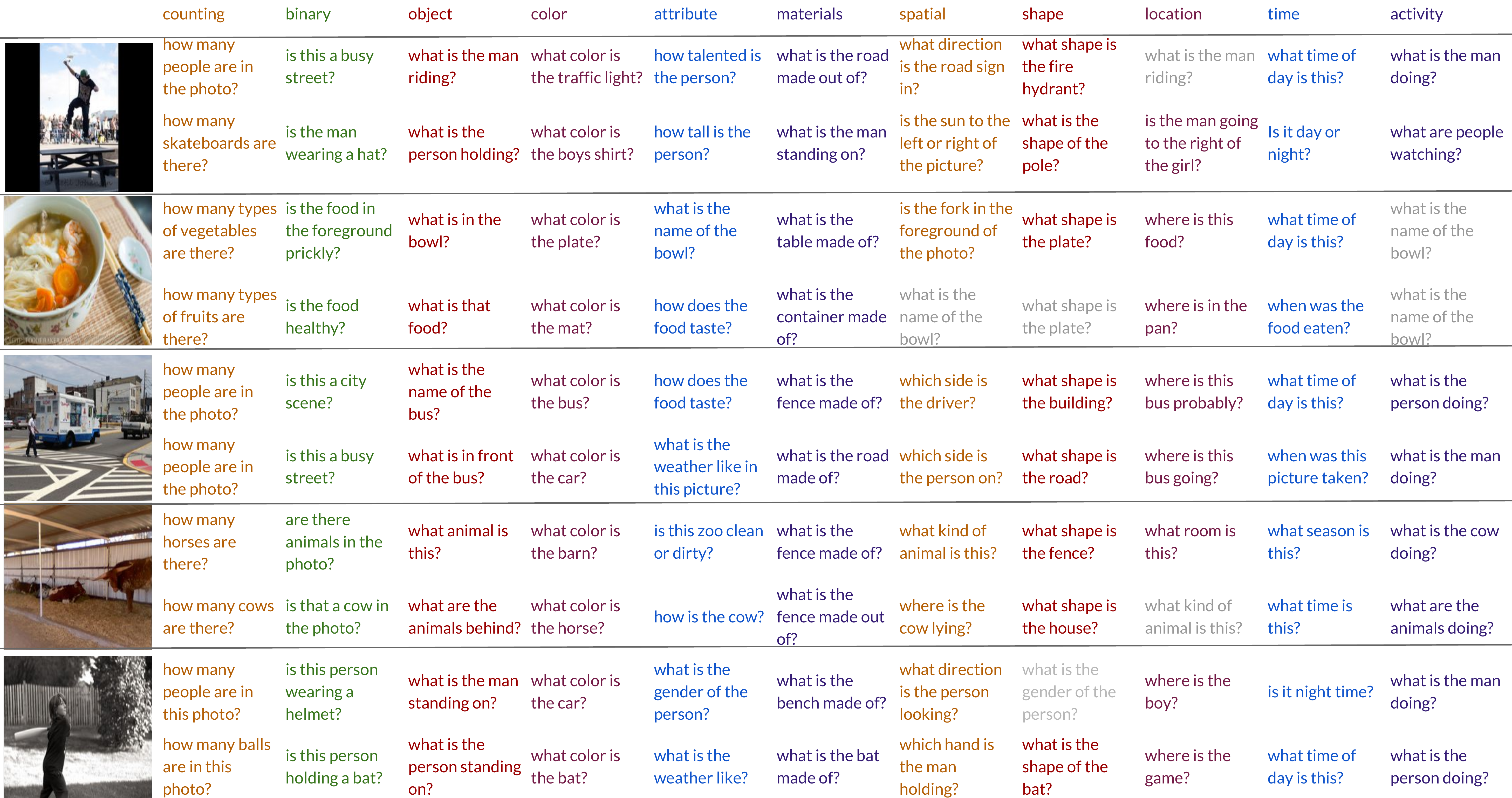}
\caption{Example questions generated for a set of images and answer categories. Incorrect questions are shown in grey and occur when no relevant question van be generated for a given image and answer category.}
    \label{fig:qualitative_results}
\end{figure*}

\noindent\textbf{Baselines.} We adapt a series of past CNN-RNN models to accept answer or answer types when generating questions. The first model \texttt{IA2Q} is a supervised, non-variational model that takes an image and answer as input and generates a question~\cite{wang2017joint}. This model is reminiscent of the VQA models often used to answer questions~\cite{wang2017joint}, except the answers are now inputs and the questions outputs~\cite{antol2015vqa,zhu2016visual7w}. Next, \texttt{V-IA2Q} is a variational version of \texttt{IA2Q}, which embeds the answer and question to a latent space before generating the question~\cite{jain2017creativity}. We also train versions of these models that accept the answer categories instead of the answer: \texttt{IC2Q} and \texttt{V-IC2Q}. When generating from a variational model, we set $z = \mu_z$ or $t=\mu_t$ to keep its outputs consistent for all measures except diversity.

We refer to our full model as \texttt{Ours} and can generate questions from either the answer latent space $z$ or the category latent space $t$. We perform ablations on this model by removing specific components. \texttt{Ours w/o A} doesn't maximize mutual information with respect to the expected answer but can also generate questions from both the $z$ and $t$ spaces. \texttt{Ours w/o C} doesn't include the $t$- space and can only generate questions from answers. Finally, \texttt{Ours w/o AC} doesn't train with the reconstruction loss nor does it have a second latent space $t$. Our evaluations  empirically demonstrate how these ablations justify our model designs. 

\subsection{Mutual information maximization}
We check whether our model improves the mutual information retained in the latent space with the input answer. We freeze the weights of a trained model and embed input images, answers and categories into the latent $z$ or $t$-space, depending on the model. We train a simple $3$-layer MLP that attempts to classify the latent code as either one of the $15$ answer categories or as one of the $500$ answers. We evaluate our model on the test set with a random chance of $6.67\%$ and $0.20\%$, respectively. Table~\ref{tab:mi} shows that the baseline models do a poor job of actually remembering the answer or category, justifying the need for a mutual information maximization approach. Since these models are unable to retain information about the input answers, it also explains why they often generate safe, generic, uninformative questions. Since our model can embed into both the $z$ as well as the $t$ space, we report how well these two spaces retain information. We find that \texttt{Ours} retains near perfect information about the input answer category with an accuracy of $99.02\%$ from $t$-space and $32.44\%$ from the $z$-space. We find that when trained without the $t$-space, \texttt{Ours w/o C} retains more information as it no longer has to constrain the $z$-space to regularize answers of the same category. We also visualize a TSNE~\cite{maaten2008visualizing} representation of the two spaces in Figure~\ref{fig:tsne}. Models that don't reconstruct the answer (e.g.~in \texttt{Ours w/o A}, \texttt{Ours w/o AC} or any of the baselines) show visually inseparable categories.

\subsection{Generating questions given the answers}
Since our model can produce questions from both answers as well as answer categories, we evaluate both scenarios individually. The language modeling section in Table~\ref{tab:mi} showcases how the various models perform when generating questions from the $z$-space, i.e.~generating questions from answers. We find that \texttt{Ours w/o C} performs the best over all the baselines and across all ablations of our model. This is likely because the latent space has more capacity when it is not also being regularized by the $t$-space. We find that \texttt{Ours w/o A} performs $~6$ METEOR points worse than \texttt{Ours} and \texttt{Ours w/o C} implying that forcing the model to reconstruct the answer does improve the quality of questions generated to better match the ground truth.    

\begin{table}[t]
    \caption{Diversity measures across different answer categories. We report the generation strength, percent of unique questions generated normalized by number of unique ground truth questions and generation inventiveness, percent of unique questions generated unseen during training. All questions were generated from the $t$-space of our model for a fair comparison with \texttt{V-IC2Q}.}
    \centering
    \small
    \label{tab:diversity}
    \begin{tabular}{l@{\hspace{0.3cm}} ccc@{\hspace{0.1cm}}cc}
    & \multicolumn{2}{c}{V-IC2Q} & & \multicolumn{2}{c}{Ours}\\
    \cmidrule{2-3}\cmidrule{5-6}
     & Strength & Inventive & & Strength & Inventive\\
    \midrule
    counting & 15.77 & 30.91 & & \textbf{26.06} & \textbf{41.30}\\ 
    binary & 18.15 & 41.95 & & \textbf{28.85} & \textbf{54.50}\\
    object & 11.27 & 34.84 & & \textbf{24.19} & \textbf{43.20}\\ 
    color & 4.03 & 13.03 & & \textbf{17.12} & \textbf{23.65}\\ 
    attribute & 37.76 & 41.09 & & \textbf{46.10}& \textbf{52.03} \\ 
    materials & 36.13 & 31.13 & & \textbf{45.75}& \textbf{40.72}\\ 
    spatial & 61.12 & 62.54 & & \textbf{70.17} & \textbf{68.18}\\
    food  & 21.81 & 20.38 & & \textbf{33.37}& \textbf{31.19}\\
    shape & 35.51 & 44.03 & & \textbf{45.81}& \textbf{55.65}\\
    location & 34.68 & 18.11 & & \textbf{45.25}& \textbf{27.22} \\
    people & 22.58 & 17.38 & & \textbf{36.20}& \textbf{31.29}\\
    time & 25.58 & 15.51 & & \textbf{34.43}& \textbf{25.30}\\
    activity & 7.45 & 13.23 & & \textbf{21.32}& \textbf{26.53}\\
    \hline
    Overall & 12.97 & 38.32 & & \textbf{26.06}& \textbf{52.11}\\
    \end{tabular}
\end{table}

\subsection{Generating questions with answer types}
The lower half of Table~\ref{tab:mi} evaluates how well our model and the baselines perform when generating questions in the absence of the actual answer and only in the presense of the answer categories. We find that overall, all the language metrics are slightly lower than when the questions were generated from the $z$-space. This is expected as now the questions need to be generated with only the answer category encoded in the $t$-space without knowing exactly what the answer is. Therefore, the models are penalized for asking an ``object'' question about the ``horse'' when the answer expects the question to focus on the ``saddle'' instead. We also qualitatively sample and report a random set of questions generated by our model in Figure~\ref{fig:qualitative_results}. We see that our model often uses concepts in the image to ground the questions. It asks specific questions like ``what is the bat made of?'' or ``is the man going to the right of the girl?''. However, there are categories like ``time'' that have a low diversity of training questions and result in the inevitable ``what time of day is this?'' question. The qualitative errors we have observed often occur when the model is forced to ask a question about a category that is not present in the image; it is hard to ask about ``food'' when no food is present.

\subsection{Measuring diversity of questions}
For all the $177K$ images in our test set, we generated one question per answer category, resulting in a total of $~2M$ questions. We report diversity in Table~\ref{tab:diversity} using two existing metrics: (1) Strength of generation: the percentage of unique generated questions normalized by the number of unique ground truth questions and (2) Inventiveness of generation: the percentage of unique questions unseen during training normalized by all unique questions generated. We compare our model with the baseline \texttt{V-IC2Q} which does not reconstruct the answer or the image. We find that our method results in more diverse set of questions across most categories. 
Questions asking for ``shape'' and ``materials'' tend to generate the most unseen questions as the model learns to generate questions like ``what [shape/material] is the \_\_\_\_ [made out of]?'' and injects objects in the given image into the missing blank. Answers agnostic to the image contents, such as ``time'', result in the fewest number of novel questions. 

\subsection{Real world deployment of our model}
To examine our model in a real world deployment, we generated $1$ question each for $1000$ images with hashtags \texttt{\#food}, \texttt{\#nature}, \texttt{\#sports}, \texttt{\#fashion} scraped from online public social media posts. Since our model needs an input answer category to ask a question, we trained a simple ResNet18 CNN~\cite{he2016deep} on the VQA images to output one of $4$ categories (see Table~\ref{tab:deployment}). We generated answer categories using the CNN and fed it into our model to generate the questions. The questions were sent to two crowd workers: one answered the question and the other reported the relevance of the question with the image and the answer with the answer category. We found all the questions asked by both \texttt{Ours} and \texttt{V-IC2Q} to be relevant to the image while $97.2\%$ and $56.8\%$ were relevant to the answer category. Our methods questions led to more unseen concepts.

\begin{table}[t]
    \centering
    \small
    \caption{We categorized $1000$ images into one of the answer categories, generated questions and asked crowd workers to answer them. We report the number of questions asked per category and the number of new concepts discovered by our model versus a baseline. We also show examples of new discovered concepts.}
    \label{tab:deployment}
    \begin{tabular}{l@{\hspace{0.3cm}} c@{\hspace{0.3cm}} c@{\hspace{0.3cm}} c@{\hspace{0.3cm}} l}
        Category & Questions & \texttt{V-IC2Q} & \texttt{Ours} & Examples \\ \midrule
        object & 411 & 10 & 80 & blackthorns, robins\\
        attributes & 205 & 8 & 40 & desecrated, crowned \\
        colors & 164 & 12 & 17 & burgandy, Alabaster \\
        materials & 220 & 4 & 8 & polyester, spandex
    \end{tabular}
\end{table}

\section{Conclusion}

We believe that visual question generation should be a task that is aimed at extracting specific categories of concepts from an image. We define a good question to be one that is not only relevant to the image but is also designed to expect a specific answer category. We build Information Maximizing Visual Question Generator that maximizes the mutual information between the generated question, the input image and the expected answer. We extend this model to overcome technical challenges associated with maximizing mutual information with discrete tokens and collapsing posterior while also allowing it to generate questions when the expected answer is absent. We analyze the questions using language modeling, diversity, relevance and mutual information metrics. We further show that through a real world deployment of this system, it can discover new concepts.

\paragraph{Acknowledgements.} We thank Justin Johnson, Andrey Kurenkov, Apoorva Dornadula and Vincent Chen for their helpful comments and edits. This work was partially funded by the Brown Institute of Media Innovation and by Toyota Research Institute (``TRI'') but this article solely reflects the opinions and conclusions of its authors and not TRI or any other Toyota entity.

{\small
\bibliographystyle{ieee}
\bibliography{references}
}

\end{document}